\documentclass{article}
\usepackage{graphicx}

\usepackage[final]{nips_2016}

\usepackage[utf8]{inputenc} 
\usepackage[T1]{fontenc}    
\usepackage{hyperref}       
\usepackage{url}            
\usepackage{booktabs}       
\usepackage{amsfonts}       
\usepackage{nicefrac}       
\usepackage{microtype}      

\title{End-to-End Pedestrian Collision Warning System based on a Convolutional Neural Network\\ with Semantic Segmentation}

\author{
  Heechul~Jung\\
  DGIST\\
  Daegu, Republic of Korea \\
  \texttt{heechul@dgist.ac.kr} \\
  \And
  Min-Kook~Choi\\
  DGIST\\
  Daegu, Republic of Korea \\
  \texttt{mkchoi@dgist.ac.kr} \\
  \And
  Kwon~Soon\\
  DGIST\\
  Daegu, Republic of Korea \\
  \texttt{soonyk@dgist.ac.kr} \\
  \And
  Woo Young~Jung\\
  DGIST\\
  Daegu, Republic of Korea \\
  \texttt{wyjung@dgist.ac.kr} \\
}

\begin{document}

\maketitle

\begin{abstract}
Traditional pedestrian collision warning systems sometimes raise alarms even when there is no danger (e.g., when all pedestrians are walking on the sidewalk). These false alarms can make it difficult for drivers to concentrate on their driving. In this paper, we propose a novel framework for an end-to-end pedestrian collision warning system based on a convolutional neural network. Semantic segmentation information is used to train the convolutional neural network and two loss functions, such as cross entropy and Euclidean losses, are minimized. Finally, we demonstrate the effectiveness of our method in reducing false alarms and increasing warning accuracy compared to a traditional histogram of oriented gradients (HoG)-based system.
\end{abstract}

\section{Introduction}

Advanced driver assistance systems (ADAS) are important for the prevention of vehicle accidents. ADAS contain several components, such as forward collision warning (FCW), lane departure warning (LDW), and pedestrian collision warning (PCW) systems. The PCW system is especially helpful for preventing serious damage and casualties, and as a result, many researchers and developers have tried to improve this system [7]. PCW systems usually rely on pedestrian detection and have an unfortunate tendency to produce false alarms in safe situations.

\begin{figure}[h]
\label{fig:problem}
\centering{\includegraphics[width=12cm]{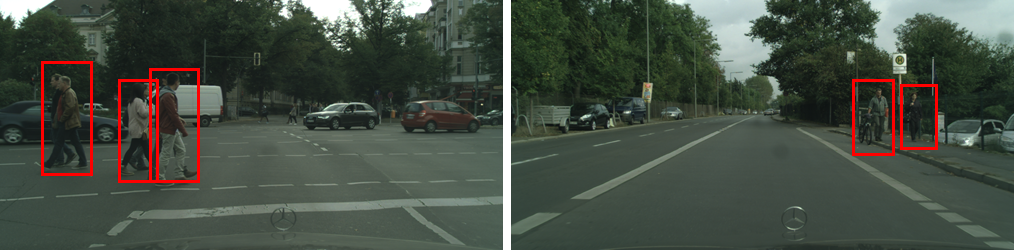}}
\caption{\textbf{Limitations of traditional PCW systems.} The image on the left shows a dangerous situation, in which drivers must be warned. The image on the right shows a safe situation for which a warning is not required.}
\end{figure}

For example, as shown in the image on the right side of Fig. \ref{fig:problem}, if pedestrians are walking on the sidewalk, then an alarm is not necessary. However, traditional pedestrian detection-based PCW systems produce alerts in this situation. It is not trivial to develop a system only using hand-crafted features because it requires several complex steps, including pedestrian detection and scene recognition. In this study, using an end-to-end framework based on a convolutional neural network (CNN), we build a system that solves this problem. 

Our contributions in this paper are summarized as follows:
\begin{itemize}
\item We propose a novel framework for a PCW system composed of an end-to-end CNN-based learning algorithm.
\item We show performance improvements of the proposed PCW system, which are achieved using semantic information from images.
\end{itemize}

There are two main advantages to our CNN-based PCW system. First, our system is effective in reducing false alarms compared to traditional pedestrian-detection-based PCW systems. Second, unlike traditional methods, our system can give warning alarms in response to cyclists as well as to pedestrians.

\section{End-to-End PCW System}
\begin{figure}[t]
\centering{\includegraphics[width=12cm]{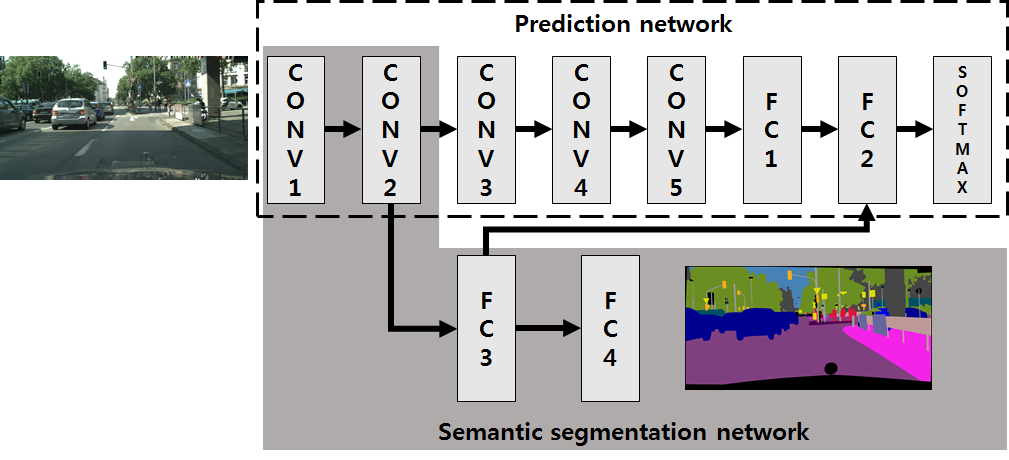}}
\caption{\textbf{Proposed CNN architecture for the PCW system.} The network in the box with dotted lines is a prediction network and the network in the gray-filled box represents a semantic segmentation network.}
\label{fig:arch}
\end{figure}

\begin{figure}[b]
\centering{\includegraphics[width=12cm]{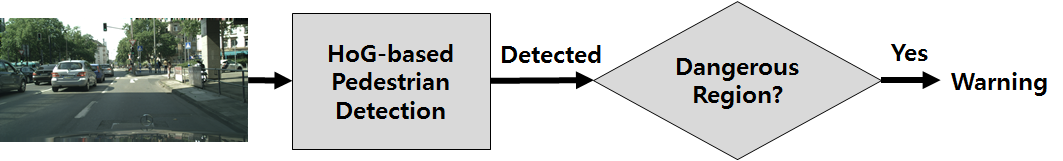}}
\caption{\textbf{Structure of a traditional HoG-based PCW system.} The baseline method has two stages: pedestrian detection and warning decision.}
\label{fig:hog}
\end{figure}

Fig. \ref{fig:arch} shows the proposed architecture of the CNN for our PCW system. We use five convolutional layers (CONV1 $\sim$ CONV5) and four fully connected layers (FC1 $\sim$ FC4). Unlike traditional approaches, which have distinct pedestrian detection and warning decision stages, as shown in Fig. \ref{fig:hog}, our PCW system does not include a pedestrian detection stage. In other words, our system predicts whether the situation is dangerous directly from the $M \times N$ raw input image. This makes our system more accurate than traditional systems, since the varying appearance of pedestrians causes the pedestrian detection stage to be imperfect.

Our network is a combination of two networks, one responsible for prediction and the other for semantic segmentation, as shown in Fig. \ref{fig:arch}. The prediction network determines whether the input image shows a warning situation, which is a binary classification problem. The semantic segmentation network segments the input image and extracts useful semantic information to feed into prediction network. These two networks are simultaneously trained by minimizing the loss function as follows:
\begin{equation}
\mathcal{L}_{t} = \mathcal{L}_c + \lambda\mathcal{L}_e,
\label{eq:total_loss}
\end{equation}
where $\mathcal{L}_t$, $\mathcal{L}_c$, and $\mathcal{L}_e$ are the total, cross-entropy, and Euclidean loss functions, respectively. $\lambda$ is a tuning parameter, which we set to $10^{-3}$ in order to adjust the scale between the two loss values.
The cross-entropy loss $\mathcal{L}_c$ is defined as follows:
\begin{equation}
\mathcal{L}_{c} = -\frac{1}{B}\sum_{i=1}^{B}\sum_{j=1}^{C}t_{ij} \log (o_{ij}^{(p)}),
\end{equation}
where $B$ is the total number of data samples in a batch. $t_{ij}$ is the $j$-th value of the ground truth label for the $i$-th data sample of the current training batch, and $o_{ij}^{(p)}$ is the $j$-th softmax output value of the prediction network  for the $i$-th data sample. $C$ is the total number of classes; in this case, $C=2$. This loss function helps the network to predict the situation correctly. The Euclidean loss function $\mathcal{L}_e$ for the semantic segmentation network is defined as follows:
\begin{equation}
\mathcal{L}_{e} = \frac{1}{2B} \sum_{i=1}^{B} ||\mathbf{o}^{(s)}_i-\mathbf{s}_i||^2_2,
\end{equation}
where $\mathbf{o}_i^{(s)} \in {\rm I\!R^{M\times N}}$ is an output vector of the last FC layer (FC4) of the semantic segmentation network for the $i$-th data sample of a batch. $\mathbf{s}_i \in {\rm I\!R^{M\times N}}$ is a vectorized form of the ground truth segmentation image for the $i$-th data sample of a batch. Using this loss function, the semantic segmentation network learns how to segment input image semantically.

These two networks share their two low level layers, CONV1 and CONV2, as shown in Fig. \ref{fig:arch}. This design is efficient, since these lower layers produce common features such as edges or blobs, allowing us to reduce the total number of learnable parameters. The output of the two networks is integrated at the FC2 layer. The high level features of FC1 and FC3 are concatenated, and the features are used as an input to the FC2 layer. Unlike the features extracted by FC1, the features extracted by FC3 represent semantic features of the input image.
The semantic features can detect and classify objects implicitly, so we expect that the semantic features will be helpful for inferring dangerous situations. We do not use the output extracted by the FC4 layer because the dimensionality of FC4 features is too large: 2048 for FC3 compared to 131,072 for FC4. This huge number of dimensions requires many weight connections at the FC2 layer that can cause over-fitting.

\section{Balancing Training Data}
It is difficult to train deep neural networks if the training data are imbalanced; that is, if the amount of data varies significantly between classes [4]. This results in imbalanced loss values between classes and failure in training the CNN. In our case, the number of images with no warning case is five times greater than the number of images showing a warning case. To resolve this problem, we copy the warning case training images in order to generate the same number of images as the non-warning case.

\section{Experimental Results}
\subsection{CNN Parameter Details}
The detailed parameter settings for our network, which is similar to AlexNet [2], are as follows: The input image size is $512\times256$, with three channels representing RGB values. For the prediction network, we use CONV(11, 96, 4) - ReLU - MaxPool(3, 2) - CONV(5, 256, 1) - ReLU - MaxPool(3, 2) - CONV(3, 384, 1) - ReLU - CONV(3, 256, 1) - ReLU - MaxPool(3, 2) - FC(256) - ReLU - FC(256) - ReLU - Softmax(2), where each value in brackets means CONV(kernel size, the number of channels, stride), MaxPool(kernel size, stride), and FC(the number of output node). For the semantic segmentation network, we use CONV(11, 96, 4) - ReLU - MaxPool(3, 2) - CONV(5, 256, 1) - ReLU - MaxPool(3, 2) - FC(2048) - FC(131072).

The size of mini-batch is of $128$, and the value of learning rate is of $0.001$. Further, we set the value of weight decay to $0.0001$, and the total iteration number is of $2000$. For the weight initialization of whole weight layers, we used the method as described in [5]. 

\subsection{Dataset}
We use a cityscape dataset taken from urban environments to evaluate our method [3]. The dataset includes various objects, such as vehicles, pedestrians, and cyclists. Furthermore, each image is densely or sparsely annotated for semantic segmentation tasks. In our experiments, we use densely annotated data consisting of $2975$ training images and $1525$ test images. Unfortunately, the dataset does not provide the ground truth data for PCW, so we have annotated each image manually (0: no alarm, 1: warning).

\subsection{Comparison to an HoG-based PCW System }
We built a baseline algorithm, shown in Fig. \ref{fig:hog}, that includes an HoG-based pedestrian detection method, for comparison with our method [1]. The rule for making a determination about the danger of a situation was: If pedestrians exist in a dangerous region, determined manually by setting a region of interest from (128, 0) to (383, 255) in the input image, then a warning alarm should be provided to the driver. (The size of the input image is $512 \times 256$.). This assumption is natural since sidewalk regions can be significantly reduced in these images.

\subsection{Results}
\begin{figure}[b!]
\centering{\includegraphics[width=6.5cm]{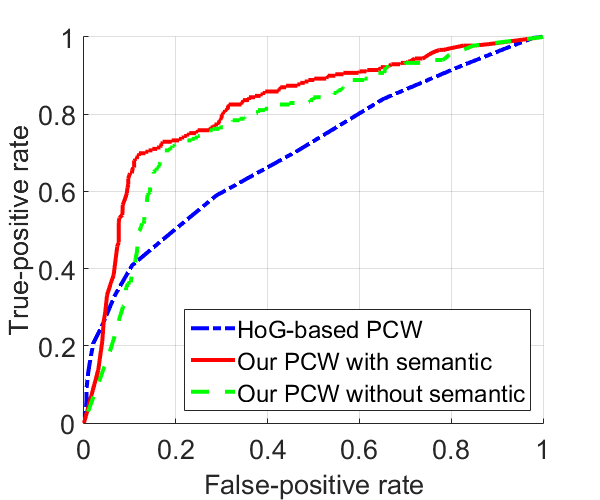}\includegraphics[width=6.5cm]{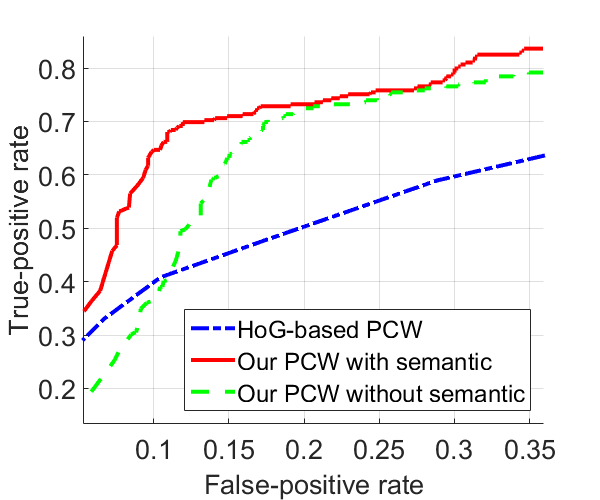}}
\caption{\textbf{ROC curves for each method.} The graph on the left shows the original ROC curve; the graph on the right shows a magnified version of the ROC curve at false positive rates in the range of $[0,~0.35]$.} 
\label{fig:roc}
\end{figure}

Fig. \ref{fig:roc} shows the receiver operating characteristic (ROC) curves for each method, showing the rates of both true and false positives. The blue line represents the HoG-based algorithm and the red and green lines denote our proposed method with and without a semantic segmentation network, respectively. The true positive rate of the HoG-based algorithm is better than that of the other methods when the false positive rate is under 0.05; however, our proposed methods are superior to the HoG-based algorithm in other cases. In particular, our proposed method with a semantic segmentation network shows the best performance among the three approaches. 

Table \ref{table} shows the accuracy (true positive rate) of each method at a 15\% false positive rate. Our proposed method without the semantic segmentation network shows about a $19\%$ improvement compared to the HoG-based algorithm. Furthermore, with the semantic segmentation network, performance was significantly improved, by $26\%$.

\begin{table}[h!]
\renewcommand{\arraystretch}{1.2}
\centering
\caption{Warning accuracies at a 15\% false positive rate.}
\label{table}
\begin{tabular}{|c|c|c|c|}
\hline
&  Baseline & Proposed & Proposed \\
&  (HoG-based) & (without semantic & (with semantic\\
&   & segmentation) & segmentation)\\
\hline
\hline
Accuracy & 45\% & 63.94\% & \textbf{71}\% \\
\hline
\end{tabular}
\end{table}

\begin{figure}[h]
\centering{\includegraphics[width=12cm]{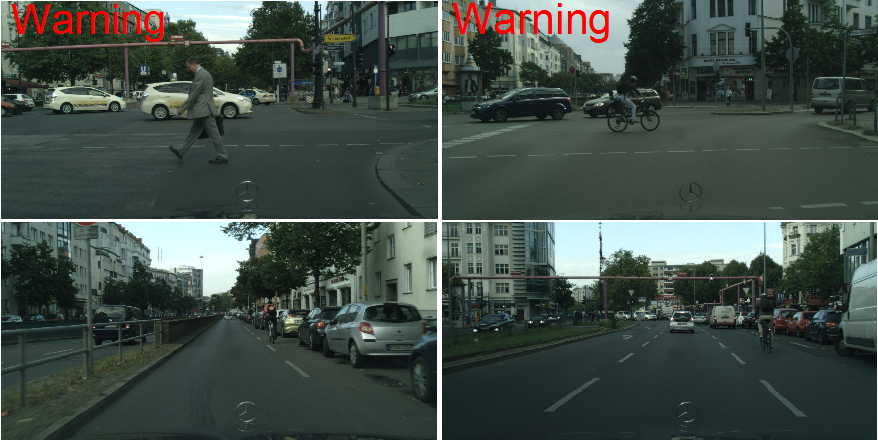}}
\caption{\textbf{Qualitative results.} The first row shows two result images for warning cases produced by the proposed method. The second row shows safe cases predicted by the proposed method.}
\label{fig:qual}
\end{figure}

Fig. \ref{fig:qual} shows qualitative results extracted from our proposed PCW system with a semantic segmentation network. Surprisingly, our system was aware of both pedestrians and cyclists. Additionally, our system did not raise an alarm when there was no risk of collision with pedestrians or cyclists. This is a significant benefit of the use of our PCW system.

\section{Conclusion}
We have proposed an end-to-end PCW system based on a CNN. The usefulness of traditional systems, based on pedestrian detection, is limited by their false alarm rate. Our system, however, is effective in reducing false alarms and improves system accuracy. Our model combines two networks that individually perform  prediction and semantic segmentation tasks, and it was helpful in improving our proposed PCW system. One of the main contributions of this paper is a demonstration of the feasibility of an end-to-end PCW system. Our system could potentially be improved by replacing the deep neural network architecture to a more recent architecture, for example, a deep residual network [6]. Additionally, we believe that our proposed framework can be applied to other ADAS, such as LDW and FCW systems. 

\section{Acknowledgement}
This work was supported by the DGIST R\&D Program of the Ministry of Science of Korea (16-FA-07).

\section*{References}
[1] Dalal, N. and Bill T. `Histograms of oriented gradients for human detection' \textit{CVPR'05}. Vol. 1. IEEE, 2005.

[2] Krizhevsky, A., Sutskever, I., and Hinton, G. E. `Imagenet classification with deep convolutional neural networks' \textit{NIPS} (pp. 1097-1105), 2012.

[3] Cordts, M., Omran, M., Ramos, S., Rehfeld, T., Enzweiler, M., Benenson, R., and Schiele, B. `The cityscapes dataset for semantic urban scene understanding' \textit{arXiv preprint} arXiv:1604.01685, 2016.

[4] Masko, D., and Hensman, P. `The impact of imbalanced training data for convolutional neural networks' 2015.

[5] He, K., Zhang, X., Ren, S., and Sun, J. `Delving deep into rectifiers: Surpassing human-level performance on imagenet classification' \textit{ICCV} (pp. 1026-1034), 2015.

[6] He, K., Zhang, X., Ren, S., and Sun, J. `Deep residual learning for image recognition' \textit{arXiv preprint} arXiv:1512.03385, 2015.

[7] Geronimo, D., Lopez, A. M., Sappa, A. D., and Graf, T. `Survey of pedestrian detection for advanced driver assistance systems' \textit{IEEE transactions on pattern analysis and machine intelligence}, \textit{32(7)}, 1239-1258, 2010.

\end{document}